\documentclass{article}
\usepackage{spconf,amsmath,graphicx}

\usepackage{enumitem}
\setlist{nosep, leftmargin=14pt}

\usepackage{mwe} % to get dummy images
\usepackage{multirow}
\usepackage{graphicx} 
\usepackage{booktabs}
\title{Uncertainty-Aware Retinal Vessel Segmentation via Ensemble Distillation}
\name{
Jeremiah Fadugba$^{\star \dagger}$ \quad 
Petru Manescu$^{\ddagger}$  \quad 
Bolanle Oladejo$^{\star}$ \quad 
Delmiro Fernandez-Reyes$^{\ddagger} \quad$
Philipp Berens$^{\ast}$ }

\address{$^{\star}$ University of Ibadan, Nigeria. \\
$^{\dagger}$ African Institute for Mathematical Sciences, Rwanda.\\
$^{\dagger}$ University College London, UK.\\
$^{\ast}$ Hertie Institute for Brain Health, T\"uebingen, Germany}

\begin{document}
%\ninept
%
\maketitle
\begin{abstract}
% Uncertainty estimation is critical for reliable medical image segmentation, particularly in retinal vessel analysis, where accurate predictions are essential for diagnostic applications. Deep Ensembles, where multiple networks are trained individually, have been shown to be widely used for improving performance in medical image segmentation. However, the cost for both training and testing increases with the number of ensembles. In this work, we propose Ensemble Distillation as a robust alternative to commonly used uncertainty estimation techniques by distilling the knowledge of multiple ensemble models into a single model. Through extensive experiments on the DRIVE and FIVES datasets, we demonstrate that Ensemble Distillation achieves comparable performance via calibration and segmentation metrics, while significantly reducing computational complexity. These findings suggest that Ensemble distillation provides an efficient and reliable approach for uncertainty estimation in the segmentation of the retinal vessels, making it a promising tool for medical imaging applications.

Uncertainty estimation is critical for reliable medical image segmentation, particularly in retinal vessel analysis, where accurate predictions are essential for diagnostic applications. Deep Ensembles, where multiple networks are trained individually, are widely used to improve medical image segmentation performance. However, training and testing costs increase with the number of ensembles. In this work, we propose Ensemble Distillation as a robust alternative to commonly used uncertainty estimation techniques by distilling the knowledge of multiple ensemble models into a single model. Through extensive experiments on the DRIVE and FIVES datasets, we demonstrate that Ensemble Distillation achieves comparable performance via calibration and segmentation metrics, while significantly reducing computational complexity. These findings suggest that Ensemble distillation provides an efficient and reliable approach for uncertainty estimation in the segmentation of the retinal vessels, making it a promising tool for medical imaging applications.
\end{abstract}

\begin{keywords}
Ensemble Distillation, Uncertainty Estimation, Calibration, Image Segmentation
\end{keywords}
\section{Introduction}
\label{sec:intro}

% \textcolor{red}{What task are we addressing and Why is it important?}:

Retina blood vessel segmentation has become an indispensable tool offering insights for the assessment of various micro-vascular and ophthalmic diseases \cite{automorph}. Advances in deep learning have enabled for more precise and efficient analysis of retinal vessels leading to better segmentation performance \cite{FR_Unet, saunet, little-wnet, iternet, RV-GAN}. Despite the success of deep learning methods in the analysis of vessel segmentation, they still make overconfident predictions due to the models incapability of signaling when their predictions are wrong. Recently, a concerted effort has been made in modeling and estimating the uncertainty of deep neural networks in medical image segmentation and proposed to account for the uncertainty in the model predictions \cite{yang2017suggestiveannotation}. This is crucial to ensure reliable predictions and support decision-making in high-stakes applications such as healthcare \cite{gal2016dropout, kendall2017uncertainties}. Accurate uncertainty quantification can lead to improved quality prediction and decision support system \cite{seedat_mcu-net_2020, kohler2024efficiently}. By measuring the uncertainty in the predictions, we can improve the interpretability and validation of the model. This measurement provides valuable insight into the confidence level of the model's outputs.

A principled treatment of uncertainty is offered by a Bayesian account, where a prior e.g. over weights is combined with a data-dependent likelihood to obtain a posterior distribution. Bayesian inference as such is intractable for complex networks, hence, approximation methods have become a cheaper alternative \cite{gal2016dropout}. Monte-Carlo Dropout and Deep Ensembles are the leading methods for estimating uncertainties in deep neural networks, in particular for medical image segmentation. Previous research has shown that Deep-Ensembles \cite{lakshminarayanan_simple_2017} not only yield better performance in prediction and in quantifying uncertainty estimates but are also the leading methods for estimating uncertainties in medical image analysis \cite{lambert_trustworthy_2022}. However, despite the predictive power of deep ensembles, information on the diversity of individual members is lost due to averaging \cite{rame2021dicediversitydeepensembles}, and the computational requirement for training and evaluation is expensive, as several number of different networks need to be trained independently. Various strategies have been proposed to alleviate the computational cost of training multiple models. For example, SnapshotEnsemble \cite{huang_snapshot_2017} save model's weights at specific point during training creating multiple "snapshots" that can then be used for accessing uncertainty. BatchEnsemble \cite{wen_batchensemble_2020} creates a multiple mini-batches of data during training avoiding multiple independent models. Similarly, MaskEnsembles \cite{durasov2021masksembles} and Layer-Ensemble \cite{kushibar_layer_2022} avoid the multiple training by randomly masking neurons during training and training multiple parallel layers respectively. However, these methods often lose the information and diversity that the deep ensemble approach gives.

% \textcolor{red}{How do we solve these weaknesses}
In this work we present an alternative approach to deep ensemble by employing the use of Ensemble distillation for estimating uncertainty in medical image segmentation. This work draws on  alternative theory to how ensemble and knowledge distillation works \cite{allen-zhu_towards_2023}. Our main contributions are:
\begin{enumerate}
    \item An alternative to Deep Ensemble for accurately estimating uncertainty in retinal blood vessel segmentation
    \item A detailed analysis of the proposed approach on two publicly available datasets
\end{enumerate}

\begin{figure}[htbp]
    \centering
    \includegraphics[width=1\linewidth]{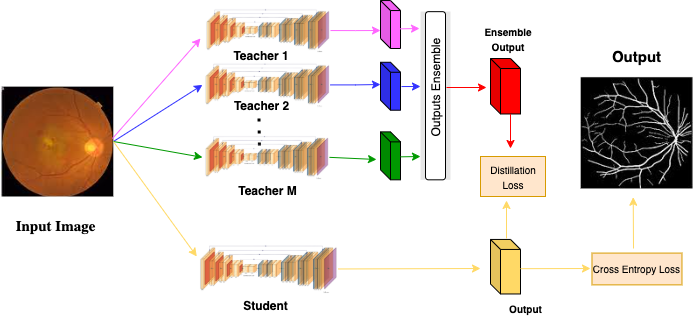}
    \caption{Ensemble Distillation Framework.}
    \label{fig:architecture}
\end{figure}

\section{Method}
% \textcolor{red}{give background to UQ, and ensemble methods and a brief intro to EnDD}
Uncertainty quantification from a Bayesian perspective gives a framework for understanding the uncertainty that exists in models and predictions. The predictive distribution is a key outcome of Bayesian inference used to make predictions about a data point \( x^* \) based on the current model and data \( D \). Given a fixed value of model parameters \( \theta \), one can learn the posterior distribution over the model parameters \(p(\theta | D)\) such that the predictive distribution for the data point can be expressed as

\[
p(y^* | x^*, D) = \int \underbrace{p(y^* | x^*, \theta)}_\text{aleatoric} \underbrace{p(\theta | D)}_\text{epistemic} d\theta.
\]

% Due to the model parameters being intractable, especially given a deep neural network, approximation techniques have been proposed (cite). One of such method is the Deep Ensemble which has been shown to have competitive predictive performance \cite{lakshminarayanan_simple_2017}. 

\textbf{Deep Ensembles}
 may be understood from a Bayesian viewpoint as an approach for approximating the posterior predictive distribution by taking the mean of the softmax outputs from each member of the ensemble.
Consider an ensemble with \( M \) models. For a given input \( x \), each model \( m \) in the ensemble provides a predictive probability distribution over the classes \( C \), denoted as \( p(y|x, \theta_m) \), where \( \theta_m \) are the parameters of model \( m \).

The method utilized by a Deep Ensemble involves synthesizing the predictions from a set of \( M \) models, each independently trained, to create a combined distribution:
The predictive distribution for each input \( x \) is the mean of the predictive distributions of the individual models in the ensemble:
   \[
   \hat{p}(y|x) = \frac{1}{M} \sum_{m=1}^{M} p(y|x, \theta_m).
   \]

In this framework, the average is taken over their predictive distributions to form the final prediction and the uncertainty estimates can be derived by \(I[p(y|x, D)]\) where \( I\) is some measures such as variance, entropy, or mutual information.

\textbf{Ensemble Distillation} extends the traditional concept of knowledge distillation by focusing on distilling the knowledge from an ensemble of models (teacher model) into a single student model \cite{hinton2015distilling, malinin2019ensemble}.
By transferring the aggregated predictive distribution of the ensembles to the single model, the goal is to train a single network parameters \( \phi \) to approximate the aggregated predictive distribution of the ensemble. 

This involves the use of Kullback-Leibler (KL) divergence, which measures the difference between two probability distributions. The objective function \( \mathcal{L}(\phi) \) for the Prior Network during training is given as :
   \[
   \mathcal{L}_{\textbf{{KL}}}(\phi) = \text{KL}(\hat{p}(y|x) \ || \ q(y|x, \phi)).
   \]
which allows the network to learn to predict the distribution of the output given an input by minimizing the loss function \( \mathcal{L}(\phi) \). This trains the network to not only predict the most likely class for each input but also capture the confidence of the ensemble about that prediction. In this work, we employed the use of similar objective function; contrastive representation distillation \cite{tian_contrastive_2022} that formulates the objective as a contrastive learning approach between the ensemble distribution and the student network. 

The contrastive loss between the student and a teacher \( i \) is formulated as:

\[
\mathcal{L}_{\text{CRD}}^i = - \log \left( \frac{\exp(\text{sim}(z^s, z^t_i) / T)}{\sum_{j=1}^{N} \exp(\text{sim}(z^s, z^t_j) / T)} \right)
\]

Where sim(·) denotes the cosine similarity between the student and teacher representations, Temperature parameter (T) is used to control the sharpness of the similarity distribution, \( (z^s, z^t_i) \) represents the positive pairs and \( N \) is the batch size.

For ensemble distillation, the total loss aggregates the contrastive losses from all teacher networks. With an ensemble of \( M \) teacher models, the final objective function becomes:
\[
\mathcal{L}_{\text{total}} = \sum_{i=1}^{M} \mathcal{L}_{\text{CRD}}^i
\]

\section{Experiment}

 We trained our network on two publicly available vessel segmentation from fundus images; FIVES dataset \cite{fives} and DRIVE dataset \cite{drive_data} which contains 800 samples and 40 samples respectively. The FIVES dataset is the largest collection of publicly available dataset for retinal vessel segmentation with robust annotation procedures. We used the official train/test split with 200 and 20 test samples for FIVES and DRIVE respectively.

We trained our segmentation model using the Full Resolution UNet model (FR-UNet)\cite{FR_Unet}, which has been shown to produce good performance in the segmentation of retinal blood vessels \cite{fadugba2024benchmarkingretinalbloodvessel}. 
All model training was carried out using PyTorch on a single NVIDIA-GeForce RTX 2080ti GPU. The models were trained for 70 epochs with a batch size of 4. We employed the Adam optimizer, incorporating a weight decay factor of $1 \times 10^{-5}$, an initial learning rate set to $1 \times 10^{-4}$, and a cosine annealing strategy for learning rate scheduling. 

We conducted a comprehensive comparison of our approach against prominent uncertainty quantification methods, including Monte Carlo Dropout (MCD) and Layer Ensemble, across multiple datasets. Our evaluation focused on both segmentation and calibration performance to ensure a holistic assessment of model reliability.

For segmentation performance, we employed two critical metrics: the Dice Similarity Coefficient (DSC) and the Matthews Correlation Coefficient (MCC). The DSC, measures the overlap between the predicted segmentation mask and the ground truth, providing insight into how well the model captures relevant structures. The MCC, evaluates the correlation between predictions and true labels. In both cases, a threshold of 0.5 was applied to binarize the predictions, ensuring consistent evaluation.

Calibration performance, essential for uncertainty quantification, was evaluated using the Expected Calibration Error (ECE) \cite{Pakdaman_Naeini_Cooper_Hauskrecht_2015}, the Brier Score (BS) \cite{Brier1950-fj}, and Negative Log-likelihood (NLL) \cite{Hastie2009-od} . ECE quantifies how well the predicted probabilities align with the true likelihood of correctness by binning the predictions and computing the gap between confidence and accuracy in each bin. The Brier Score provides a combined measure of both the accuracy and confidence of the probabilistic predictions, rewarding models that are confidently correct and penalizing those that are confidently wrong. NLL, meanwhile, captures the quality of the model's probabilistic predictions by evaluating the log-probability assigned to the true class, with lower values indicating better calibration.

\section{Results}
We analyzed multiple approaches for quantifying uncertainties, including Deep Ensemble (5 models), Monte Carlo Dropout (10 passes during test time), Layer Ensemble and compare them against our approach Ensemble Distillation with KL-Divergence and Ensemble Distillation with Contrastive representation Distillation.

\begin{table}[htbp]
\centering
\caption{Results on FIVES dataset. Values in bold indicate the best result. $\uparrow$ - higher is better, $\downarrow$ - lower is better.}
\resizebox{\columnwidth}{!}{
\begin{tabular}{cccccc}
\toprule
% & & \multicolumn{2}{c}{Segmentation Performance} & \multicolumn{3}{c}{Confidence Calibration} \\ 
 Method & DSC $\uparrow$ & MCC $\uparrow$ & ECE $\downarrow$ & BS $\downarrow$ & NLL $\downarrow$ \\ 
\midrule
 Baseline         & 0.8857 & 0.8821 & \textbf{0.0063} & 0.0107 & 0.0324 \\  
 DE               & \textbf{0.8905} & \textbf{0.8853} & 0.0068 & \textbf{0.0106} & \textbf{0.0273} \\  
 MCD              & 0.8850 & 0.8815 & 0.0075 & \textbf{0.0106} & 0.0309 \\  
 LE               & 0.8487 & 0.8432 & 0.0089 & 0.0134 & 0.0325 \\ 
 EnD-KL (ours)    & 0.8817 & 0.8770 & 0.0065 & 0.0109 & 0.0298 \\  
 EnD-CRD (ours)   & 0.8819 & 0.8797 & 0.0124 & 0.0108 & 0.0377 \\  
 
\bottomrule
\end{tabular}
\label{tab:fives_results}
}

\end{table}
\subsection{Segmentation Performance}
The results on both the FIVES and DRIVE datasets (Table \ref{tab:fives_results} and Table \ref{tab:drive_result} respectively) highlight the segmentation and calibration performance of various methods. For the FIVES dataset, the Ensemble Distillation method using KL divergence (EnD-KL) shows comparable performance to other techniques with a DSC of 0.8817, ECE (0.0065) and NLL (0.0098). Similarly, Monte Carlo Dropout (MCD) and Layer Ensemble (LE) perform competitively, but Deep Ensemble (DE) shows slightly lower calibration, indicated by a higher Brier Score (BS) of 0.0106. 
On the DRIVE dataset, the EnD-KL method again shows comparable performance to the best-performing method (MCD) with about 0.21\% gain on DSC and 0.17\% gain on the MCC.  However, EnD-CRD shows low segmentation performance across both datasets but performs moderately well in terms of calibration on the FIVES dataset, with better results than on DRIVE. Overall, Ensemble Distillation using KL divergence is an efficient and reliable method for both segmentation and calibration on these datasets.
\begin{table}[htbp]
        \centering
        \caption{Results on DRIVE dataset. Values in bold indicate best result. $\uparrow$ - higher is better, $\downarrow$ - lower is better.}
        \resizebox{\columnwidth}{!}{
            \begin{tabular}{cccccc}
            \toprule
            Method & DSC $\uparrow$ & MCC $\uparrow$ & ECE $\downarrow$ & BS $\downarrow$ & NLL $\downarrow$ \\ 
            \midrule
            % \multirow{5}{*}{DRIVE} 
            Baseline         & 0.8052 & 0.7905 & 0.0201 & 0.0264 & 0.0772 \\  
            DE               & 0.7895 & 0.7711 & 0.0259 & 0.0314 & 0.0744 \\  
            MCD              & \textbf{0.8158} & \textbf{0.8004} & \textbf{0.0163} & \textbf{0.0243} & \textbf{0.0544} \\  
            LE               & 0.7995 & 0.7825 & 0.0165 & 0.0245 & 0.0555 \\ 
            EnD-KL (ours)    & 0.8141 & 0.7990 & 0.0178 & 0.0251 & 0.0653 \\  
            EnD-CRD (ours)   & 0.7999 & 0.7853 & 0.0170 & 0.1069 & 0.1005 \\  
            \bottomrule
            \end{tabular}
        }
        \label{tab:drive_result}
\end{table}

\subsection{Calibration Analysis}
\begin{figure}
    \centering
        \includegraphics[width=1\linewidth]{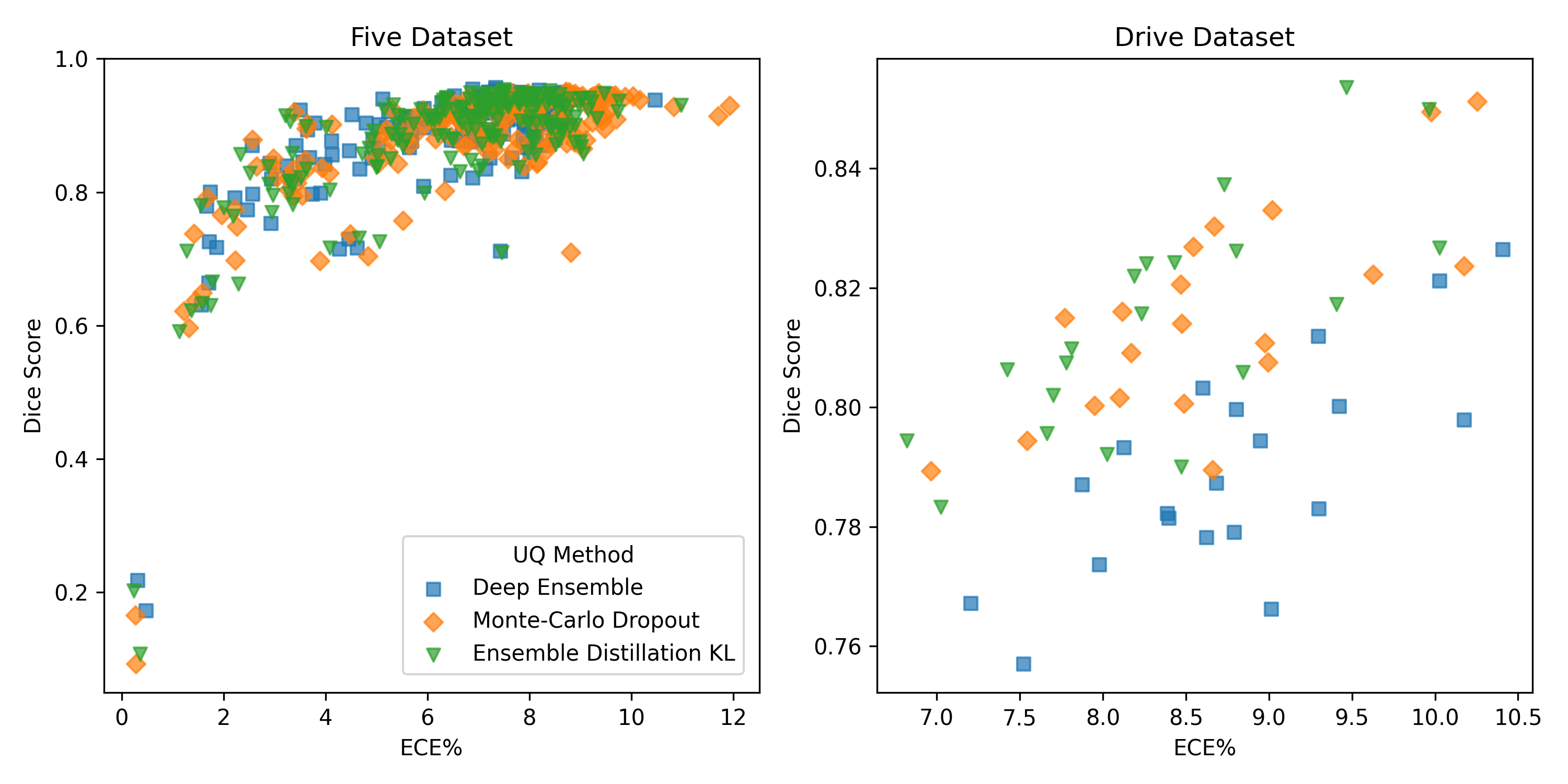}
    \caption{Comparison of ECE\% and Dice score for Various UQ method where each point represents an image in the test set. On FIVES (200 samples), there's a similar trade-offs between calibration and segmentation performance while on DRIVE (20 samples) dataset, EnD-KL achieves relatively higher Dice score with lower ECE\% than Deep Ensembles.}
    \label{fig:ece_dice}
\end{figure}
    \centering
For the DRIVE dataset, the calibration analysis compares six uncertainty techniques, revealing that the baseline model is reasonably well-calibrated but slightly overconfident in high probability bins (>0.8). Monte Carlo Dropout, Layer Ensemble, and the developed Ensemble Distillation using KL divergence show improved calibration, with Layer Ensemble being slightly under-confident in the mid-range probabilities. The Deep Ensemble method shows significant under-confidence, while Ensemble Distillation with KL divergence and CRD performs better, although CRD tends to be overconfident for probabilities above 0.6. For the FIVES dataset (200 samples), similar trends are observed. Monte Carlo Dropout and KL distillation provide the best calibration, while CRD shows overconfidence in higher probabilities and Deep Ensemble performs better than on the DRIVE dataset, yet remains under-confident at lower probabilities. Overall, Monte Carlo Dropout and Ensemble Distillation using KL divergence emerge as the most reliable techniques across both datasets.
\begin{figure}[ht]
    \centering
    \includegraphics[width=1\linewidth]{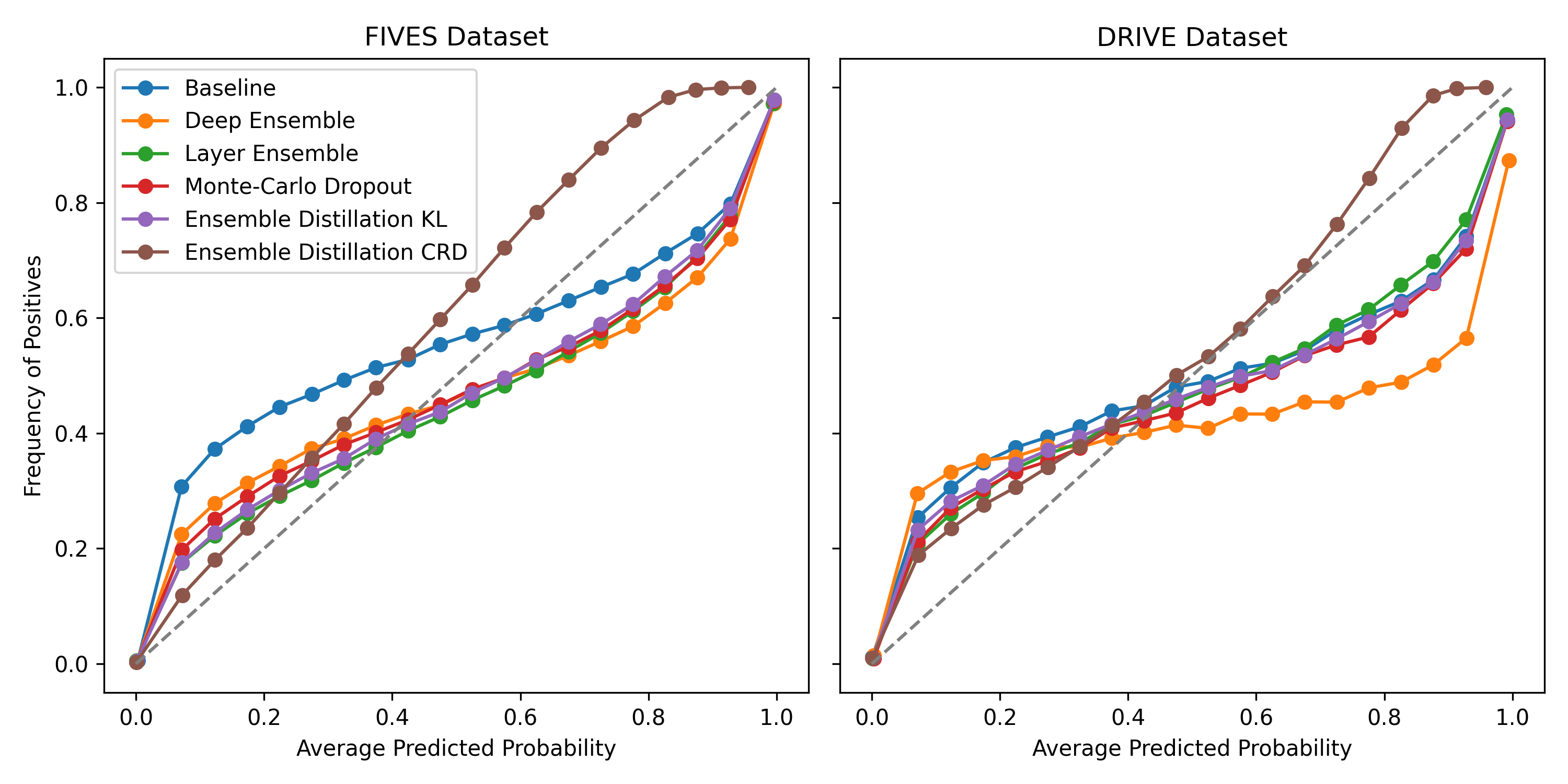}
    \caption{Reliability Diagram showing the calibration performance of different UQ methods, with the diagonal line representing perfect calibration (where predicted probabilities match the observed frequencies}
    \label{fig:reliability_diagram}
\end{figure}

\section{Conclusion}

In this work, we presented Ensemble Distillation with KL Divergence (EnD-KL) as an alternative to Deep Ensembles (DE) for uncertainty quantification in retinal vessel segmentation. Our experiments on two retinal vessel datasets demonstrated that EnD-KL achieves segmentation performance comparable to DE, as evidenced by similar Dice scores in both datasets, while maintaining robust calibration. Reliability diagrams and ECE scores confirmed that EnD-KL provides well-calibrated uncertainty estimates, closely matching the calibration levels of DE. Furthermore, EnD-KL significantly reduces computational demands by consolidating ensemble knowledge into a single model, which is particularly advantageous for deployment in resource-constrained environments. These findings establish EnD-KL as a viable, resource-efficient alternative to DE, capable of delivering accurate segmentation and reliable uncertainty estimation. Future research could expand on the application of this method to other medical imaging tasks, reinforcing its potential to enable trustworthy and accessible AI-driven diagnostics in healthcare.

\section{Acknowledgments}
\label{sec:acknowledgments}
This publication was made possible by a grant from the Carnegie Corporation of New York (provided through the African Institute for Mathematical Sciences). 
% The statements made and views expressed are solely the responsibility of the author(s).

\bibliographystyle{IEEEbib}
\bibliography{refs}

\end{document}